%% file: aaai2026.tex
\documentclass[letterpaper]{article} 
\usepackage{aaai2026}  
\usepackage{array}     
\usepackage{multirow}  
\usepackage{makecell} 

\makeatletter
\def\copyright@text{}%
\def\copyright@on{F}%
\def\copyrighttext#1{}%
\def\copyrightyear#1{}%
\def\@copyrightspace{}%
\makeatother

\usepackage{amsmath}
\usepackage{amsfonts}
\usepackage{amssymb}

\usepackage{times}  
\usepackage{helvet}  
\usepackage{courier}  
\usepackage[hyphens]{url}  
\usepackage{graphicx} 
\usepackage{natbib}  
\usepackage{caption} 
\usepackage{algorithm}
\usepackage{algorithmic}

\usepackage{newfloat}
\usepackage{listings}
\DeclareCaptionStyle{ruled}{labelfont=normalfont,labelsep=colon,strut=off} 
\floatstyle{ruled}
\newfloat{listing}{tb}{lst}{}
\floatname{listing}{Listing}
\title{DiffPCN: Latent Diffusion Model Based on Multi-view Depth Images for Point Cloud Completion}
\author{
    Zijun Li\textsuperscript{\rm 1,3}\equalcontrib,
    Hongyu Yan\textsuperscript{\rm 2}\equalcontrib,
    Shijie Li\textsuperscript{\rm 3},
    Kunming Luo\textsuperscript{\rm 2},
    Li Lu\textsuperscript{\rm 1}\thanks{Corresponding author.},
    Xulei Yang\textsuperscript{\rm 3},
    Weisi Lin\textsuperscript{\rm 4}
}
\affiliations{
    \textsuperscript{\rm 1}Sichuan University\\
    \textsuperscript{\rm 2}Hong Kong University of Science and Technology\\
    \textsuperscript{\rm 3}Institute for Infocomm Research (I2R), A*STAR, Singapore\\
    \textsuperscript{\rm 4}Nanyang Technological Unversity\\

}

\usepackage{bibentry}

\begin{document}

\maketitle

\input{sec/0_abstract}

\input{sec/1_intro}
\input{sec/2_related_work}
\input{sec/3_preliminary}

\input{sec/4_method}

\input{sec/5_experiment}

\input{sec/6_conclution}

\clearpage

\bibliography{aaai2026}

\end{document}

%% file: sec/0_abstract.tex
\begin{abstract}

Latent diffusion models (LDMs) have demonstrated remarkable generative capabilities across various low-level vision tasks. However, their potential for point cloud completion remains underexplored due to the unstructured and irregular nature of point clouds. In this work, we propose~\textbf{DiffPCN}, a novel diffusion-based coarse-to-fine framework for point cloud completion. Our approach comprises two stages: an initial stage for generating coarse point clouds, and a refinement stage that improves their quality through point denoising and upsampling. Specifically, we first project the unordered and irregular partial point cloud into structured depth images, which serve as conditions for a well-designed DepthLDM to synthesize completed multi-view depth images that are used to form coarse point clouds. In this way, our DiffPCN can yield high-quality and high-completeness coarse point clouds by leveraging LDM’s powerful generation and comprehension capabilities. Then, since LDMs inevitably introduce outliers into the generated depth maps, we design a Point Denoising Network to remove artifacts from the coarse point cloud by predicting a per‑point distance score. Finally, we devise an Association-Aware Point Upsampler, which guides the upsampling process by leveraging local association features between the input point cloud and the corresponding coarse points, further yielding a dense and high‑fidelity output. Experimental results demonstrate that our DiffPCN achieves state-of-the-art performance in geometric accuracy and shape completeness, significantly improving the robustness and consistency of point cloud completion.
\vspace{-2mm}
\end{abstract}

%% file: sec/1_intro.tex
\section{Introduction}
\begin{figure}[t]
\centering
\scalebox{0.45}{
\includegraphics[width=\textwidth]{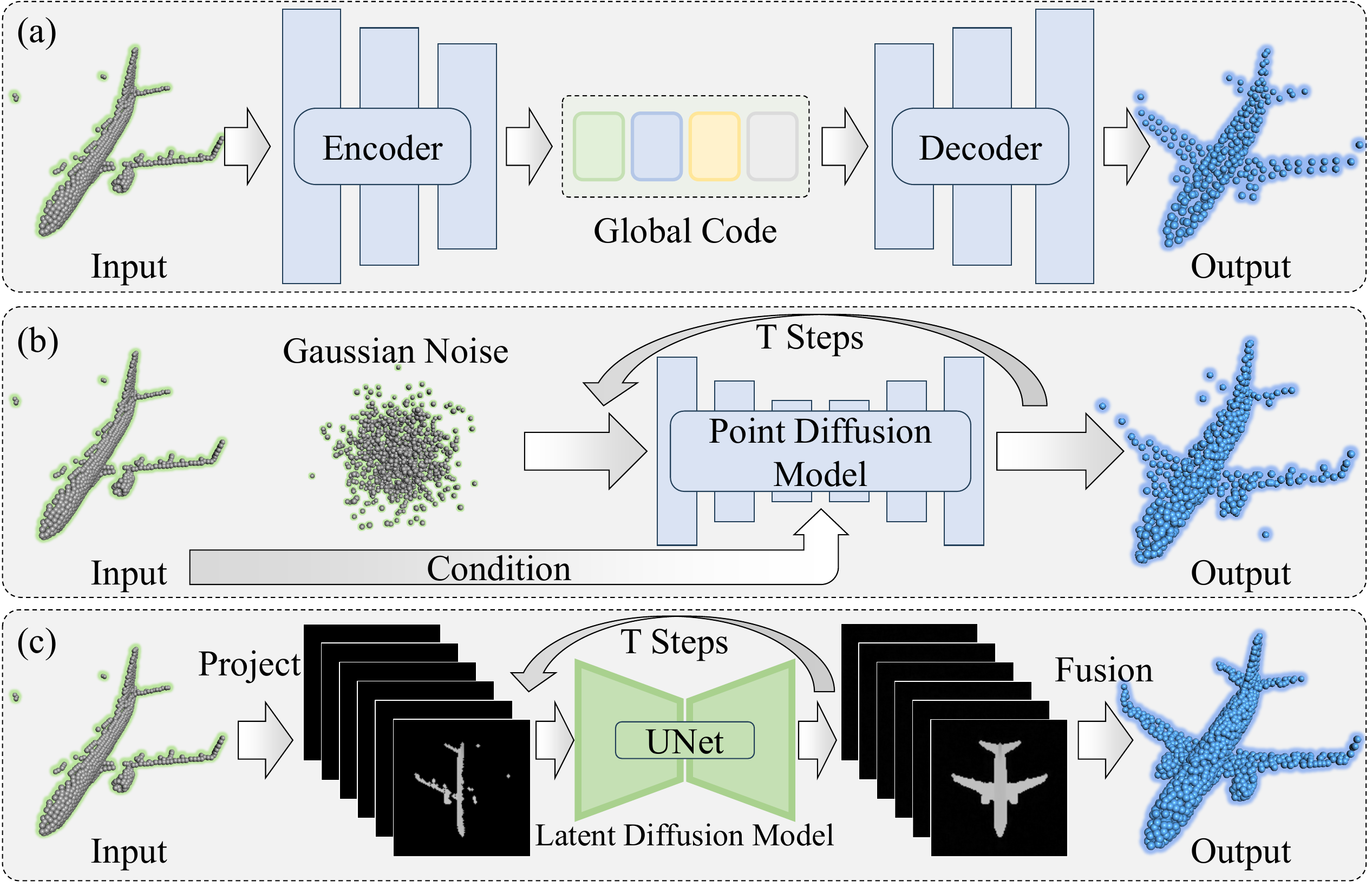}}
\caption{The first stage of mainstream point cloud completion approaches. They are (a): point based methods, (b): point diffusion based methods, and (c): our depth diffusion based method respectively. For each baseline, we render the best-performing point-count setting reported in its paper~\cite{zhu2023svdformer,lyu2021conditional}.}
\label{fig:Compare}
\vspace{-4mm}
\end{figure}

Point clouds, directly captured by sensors such as LiDAR and depth cameras, provide fundamental 3D spatial information. However, real-world point clouds are often sparse and incomplete due to occlusion, sensor noise, and limited resolution, adversely affecting downstream 3D tasks~\cite{ben20183dmfv, qiu2021dense, landrieu2018large, tang2022contrastive, rajathi2023path}. Therefore, Point cloud completion is proposed aiming to reconstruct dense, structurally coherent point clouds from partial observations, significantly improving 3D perception and reasoning.

\begin{figure*}[t]
\centering
\scalebox{0.95}{
\includegraphics[width=\textwidth]{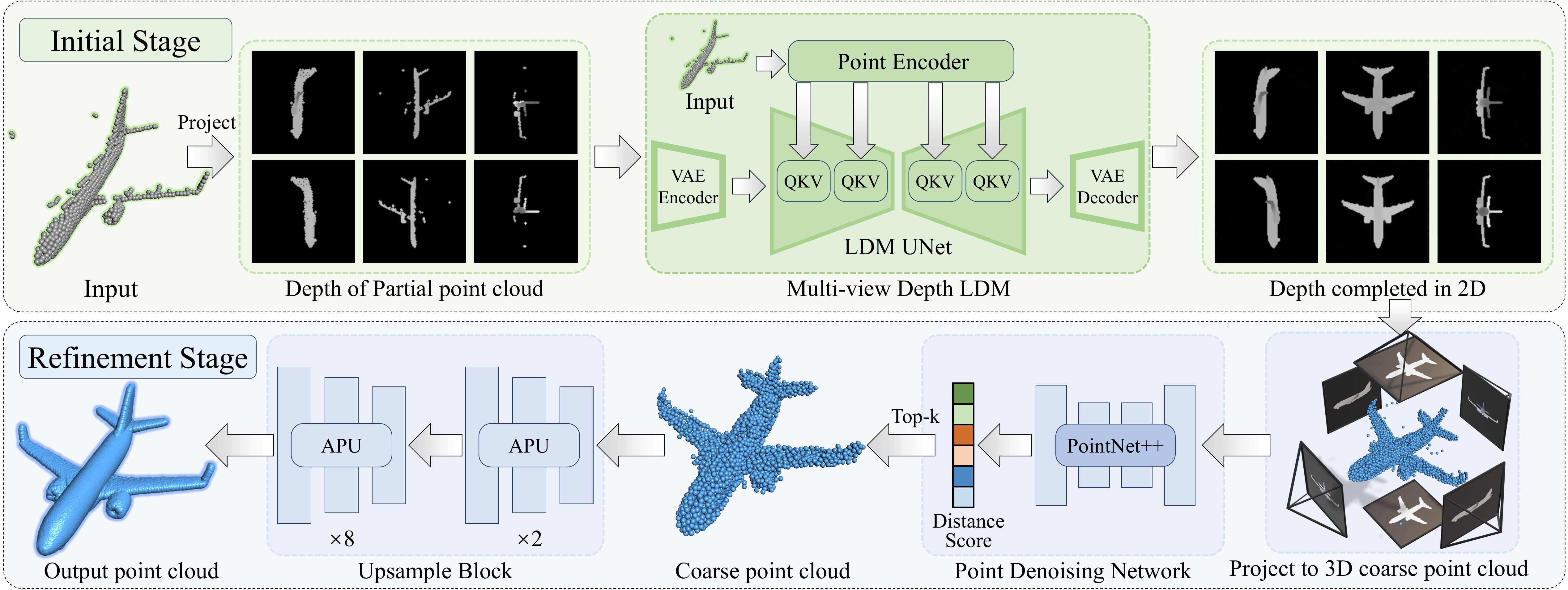}}
\caption{The overall pipeline of DiffPCN. Our DiffPCN consists of two primary stages: the initial stage and refinement stage. The initial stage aims to generate a high-quality initial point clouds from a latent diffusion model based on multi-view depth images from input point clouds. Subsequently, the refinement stage is used to refine and improve the details of the initial point clouds, which is composed of a point denoising network and two stacked association-aware upsampling networks.}
\label{fig:framework}
\vspace{-2mm}
\end{figure*}

Most existing point cloud completion approaches follow a coarse-to-fine reconstruction pipeline, where an initial coarse completion is first generated and then refined to improve structural details. Recent studies~\cite{chen2023anchorformer, zhou2022seedformer, li2023proxyformer} emphasize that the quality of initial completion crucially impacts the final reconstruction accuracy and detail. An effective and high-quality initial completion should not only be accurate but also capture fine-grained details. Earlier methods ~\cite{xiang2021snowflakenet, yu2021pointr, yan2022fbnet}, typically extract a global feature from the partial input and decode it to reconstruct a coarse point cloud, as shown in Figure~\ref{fig:Compare} (a). However, these global feature based methods inadvertently disrupt point cloud’s original spatial structure and suffer from severe information loss, leading to sparse and structurally inconsistent predictions. To address this issue, recent approaches integrate additional cues to enrich the initial generation process. For instance, SVDFormer~\cite{zhu2023svdformer} integrates depth information to provide explicit geometric constraints, and PCDreamer~\cite{wei2024pcdreamer} fuses global features of generated multi-view images and partial point clouds to predict coarse point clouds. Although these methods can improve feature representation by utilizing cross-modality information, they still rely on global features to form the initial point cloud, typically no more than 256 points, leading to a failure to capture fine-grained local structures.

Recently, diffusion models~\cite{ho2020denoising, song2020denoising, ho2022classifier} have demonstrated remarkable success in generating high-quality content. Many low-level tasks~\cite{ke2024repurposing, nichol2021glide, shiohara2024face2diffusion, kim2024stableviton} also used the diffusion model to solve their problem and achieved impressive performance. Similarly, as shown in Figure~\ref{fig:Compare} (b), some completion methods~\cite{lyu2021conditional, chen2024learning} also tried to utilize the diffusion model to directly generate coarse point clouds conditional on partial input. However, because these methods directly construct the diffusion model in the 3D space rather than the latent space, they face the efficiency problem. In addition, the irregularity and disorder of point clouds hinder the effectiveness of the diffusion model that works well in continuous representations, such as the image~\cite{rombach2022high,saharia2022palette}, video~\cite{ho2022video, khachatryan2023text2video}, and grid~\cite{li2024craftsman}. Therefore, they still fail to efficiently synthesize geometrically faithful coarse point clouds, which undermines the entire coarse-to-fine pipeline. 

To address the irregularity of raw point sets and harness the robust generalization of latent diffusion model, we introduce DepthLDM, a conditional LDM built on structured multi-view depth images as shown in Figure~\ref{fig:Compare} (c). Inspired by MVDD~\cite{wang2024mvdd}, which uses DDPM for unconditional multi-view depth images generation, we first project the unordered input into continuous multi-view depth images. Conditioned on the 2D features of multi-view incomplete depth maps and the 3D features extracted from raw 3D partial point clouds, DepthLDM can generate multi-view consistent depth images to get a high-quality coarse point cloud by leveraging cross-view attention introduced in Wonder3D~\cite{long2024wonder3d} and specialized point-aligned attention that can enhance the LDM’s structural consistency and space comprehension. Then, we form the coarse point clouds by directly projecting the depth map back to 3d space. As a result, we can yield more complete and accurate coarse point clouds than the previous methods.

Based on DepthLDM, we introduce a novel coarse-to-fine pipeline for point cloud completion, called {\bf DiffPCN}, as shown in Figure~\ref{fig:framework}. After obtaining the coarse point cloud from DepthLDM, further denosing is required to address artifacts introduced during reconstruction from generated depth images. These artifacts deviate from the object's boundaries often lead to abnormal points in coarse point cloud~\cite{ke2024repurposing}. To mitigate this issue, we introduce the Point Denoising Network (PDNet), which learns to estimate the distance score for each point in coarse point cloud. Using this estimated score, we apply a top-k algorithm to remove points with the highest distance scores, ensuring a more structured and clearer initial point cloud.

Finally, following previous coarse-to-fine pipelines~\cite{xiang2021snowflakenet, zhou2022seedformer, Rong2024CRAPCNPC}, we refine the initial outputs with our Association-Aware Point Upsampler (APU). Since the input point cloud is the sole repository of object texture and detail information, our key insight in designing the APU was to learn the local displacement association between the partial input and the coarse prediction. APU operates in two stages: first, an Association Transformer extracts high-frequency deformation cues by learning the corrective transformation between reliable partial input with its corresponding subset of the coarse prediction; Second, these learned cues are diffused into the full initial point cloud through a cross-attention layer and propagated globally via successive self-attention blocks. This association-driven upsampling yields a dense, geometrically faithful final point cloud that reconstruct both global structure and fine local details.

To validate the effectiveness of our proposed DiffPCN, we conduct extensive experiments on multiple benchmark datasets, including PCN~\cite{yuan2018pcn}, ShapeNet55/34 dataset~\cite{yu2021pointr} and MVP dataset~\cite{pan2021variational}. Our method outperforms existing state-of-the-art approaches in both quantitative metrics and qualitative results, demonstrating superior completeness and structural fidelity. Our main contributions are summarized as follows:

\begin{itemize}
\item We propose a novel framework, DiffPCN, to introduce the latent diffusion model for point cloud completion, which achieves state-of-the-art performance on multiple completion benchmarks.

\item We propose DepthLDM to transfer the 3D point cloud into the multi-view depth images to inherit the LDM's powerful generation ability to generate high-quality initial point clouds.

\item We propose a point denoising network (PDNet) to remove the outliers by computing the confidence score, as well as designing an association-aware point upsampler (APU) to obtain fine-grained complete point clouds by leveraging the association relation between partial and coarse point clouds.
\end{itemize}


%% file: sec/2_related_work.tex
\section{Related Work}
\subsection{Point-based Methods}
The introduction of PointNet and PointNet++~\cite{qi2017pointnet, qi2017pointnet++} marked a significant breakthrough in the field of point cloud processing. Based on this, a pioneering method PCN~\cite{yuan2018pcn} proposed a coarse-to-fine pipeline and a folding-based upsampling block to refine coarse point clouds. Subsequent methods ~\cite{wang2020cascaded, yang2018foldingnet, tchapmi2019topnet, pan2021variational} extended this idea to further improve performance by leveraging multi-scale local feature extractors and encoder-decoder architectures. Compared to traditional grid-based methods, PointNet-based approaches significantly reduce computational complexity and handle sparse point clouds more effectively~\cite{aoki2019pointnetlk}. However, challenges remain in capturing intricate geometries and global structures. Inspired by the powerful attention machine in transformers, PoinTr~\cite{yu2021pointr} initiated a new wave of research, which was subsequently advanced by several work~\cite{ xiang2021snowflakenet, zhu2023svdformer, yan2022fbnet}. Other methods~\cite{chen2023anchorformer, zhou2022seedformer, zhang2022point} modeled regional discrimination by learning a set of anchors or patches through self-attention. Furthermore, inspired by advances in image‐based recognition~\cite{choy20163d, wei2022learning, yang2019learning}, view‐guided completion methods~\cite{zhang2021view, aiello2022cross, zhu2023svdformer} were designed to leverage 2D CNNs or ViTs~\cite{liu2021swin} to inject the global feature of RGB or depth image into point clouds. Despite the improvements, the model's inherent limitations in understanding 3D shapes constrain its ability to generate initial point clouds with rich details and logical structures.

\subsection{Diffusion-based Methods}
As a new generation of generative models, diffusion probabilistic model~\cite{ho2020denoising} has garnered significant attention since its introduction. It leverages a Markov chain to transform the noise distribution into the data distribution. Subsequent research has explored various aspects~\cite{song2020denoising, luo2023latent, ho2022classifier}, greatly enhancing the practicality and efficiency of diffusion models. As a result, diffusion models have been successfully applied to many downstream tasks~\cite{ho2022video,saharia2022photorealistic,ke2024repurposing, nichol2022point, luo2021diffusion}, achieving excellent performance across diverse domains.

In the field of 3D point cloud completion, diffusion models have demonstrated promising potential, offering new opportunities to address the challenges of conditional shape reconstruction and semantic consistency. PDR~\cite{lyu2021conditional} is the first work to utilize diffusion model for point cloud completion. It employs DDPM~\cite{song2020denoising} and PointNet++~\cite{qi2017pointnet++} to generate a coarse completion conditioned on partial observations. PointLDM ~\cite{chen2024learning} further advances this direction by leveraging a coarse-to-fine PointVAE to refine the initial point cloud. However, due to the irregular and unordered nature of point clouds, VAE and diffusion models in the point cloud domain often struggle to achieve optimal performance. Recent work PCDreamer ~\cite{wei2024pcdreamer} harnesses large multi-view diffusion models ~\cite{shi2023mvdream, long2024wonder3d} to generate novel views of RGB images for partially observed shapes, using the feature extract from them to predict initial coordinates. While this approach effectively leverages the prior knowledge of 3D space provided by foundation model, it suffers from extremely high computational complexity and relatively low information utilization because they still follow the paradigm of using the global feature to generate a corase point cloud. In a word, existing diffusion-based methods for point cloud completion face common challenges, inefficient global feature utilization, high computational cost, and difficulty in capturing fine-grained structures from disorder point clouds, which limit their reconstruction quality and scalability.


%% file: sec/3_preliminary.tex
\section{Preliminary}

Diffusion models have emerged as powerful generative frameworks, demonstrating remarkable performance across diverse modalities such as images, audio, and 3D data~\cite{ho2020denoising, dhariwal2021diffusion, li2024craftsman}. The core idea is to progressively map a data distribution to Gaussian noise through a forward process, then learn to reverse this transformation to generate new samples from gaussian noise. Given a data sample $ x_0 $, the forward process adds Gaussian noise at each step:
\begin{equation}
    q(x_t \mid x_{t-1}) = \mathcal{N}(x_t; \sqrt{\alpha_t} x_{t-1}, (1-\alpha_t) \mathbf{I}),
\end{equation}
where $\alpha_t $ controls the noise level. After sufficiently many steps $( t = T )$, the distribution degenerated to an isotropic gaussian.
To recover $ x_0 $, a neural network is trained to approximate the reverse process, parameterized as:
\begin{equation}
    p_\theta(x_{t-1} \mid x_t, y) = \mathcal{N}(x_{t-1}; \mu_\theta(x_t, t, y), \Sigma_\theta(t)),
\end{equation}
where $y$ represents conditioning information such as class labels~\cite{ho2022classifier}, text embeddings~\cite{ramesh2022hierarchical}, or structural priors~\cite{ nichol2022point}. The model minimizes the reconstruction error under conditioning:
\begin{equation}
    \mathcal{L} = \mathbb{E}_{x_0, \epsilon, t} \Big[ \lVert \epsilon - \epsilon_\theta(x_t, t, y) \rVert_2^2 \Big],
\end{equation}
where $\epsilon$ is the added noise, and $\epsilon_\theta$ is the noise predictor which is usually UNet structure. Conditional diffusion models enhance control over generation and have shown strong performance in several tasks~\cite{ke2024repurposing, zhang2024diffusion}.

%% file: sec/4_method.tex
\section{Method}
\subsection{Overview}
In this section, we introduce DiffPCN, a coarse-to-fine diffusion-based framework for point cloud completion. As illustrated in Figure~\ref{fig:framework}, DiffPCN consists of two primary stages: the initial stage and the refinement stage. The initial stage aim to generate a coarse point cloud, while the refinement stage focuses on improving the quality of coarse point clouds.


\subsection{Initial Stage}
\label{DepthLDM}
In this stage, we aim to generate a reliable and detailed initial point cloud. To achieve this, we project the point cloud into the 2D depth image and propose a depth-based latent diffusion model {\bf DepthLDM}, as shown in the upper half of Figure~\ref{fig:framework}.

Specifically, we first train a Variational Autoencoder (VAE) on depth maps projected from ground truth point clouds. For each point cloud, we render 6 views depth images with 128 $\times$ 128 resolution by using pre-defined camera poses, including front, back, left, right, up, and down. Similar to Stable Diffusion~\cite{rombach2022high}, our VAE comprises an encoder, a KL regularization block, and a decoder. Specifically, given an input depth map $I \in \mathbb{R}^{H \times W \times 1}$, it will be encoded as a latent code $Z \in \mathbb{R}^{H^\prime \times W^\prime \times C}$ which will be further reconstructed into depth image $I^{\prime}$ by decoder:
\begin{align}
z = KL(\mathcal{E}(I)), I^{\prime} = \mathcal{D}(z)
\end{align}
Where $KL$ indicates KL regularization. $\mathcal{E}$, $\mathcal{D}$ is our encoder and decoder. We trained the depth VAE following the standard approach and define the losses as:
\begin{equation}
    \mathcal{L}_{vae} = \mathrm{MSE}\bigl(I_{\mathrm{out}}, I'\bigr) \;+\; \lambda\,L_{\mathrm{kl}}\bigl(p(z,I'), \mathcal{N}(0, I)\bigr),
\end{equation}

Then, we build a conditional diffusion model to generate depth images from noise. In practice, we observe that generating each view independently with a diffusion model leads to view inconsistency, resulting in low-quality initial point clouds. 
To address this issue and improve multi-view consistency, we adopt a cross-view attention mechanism, following Wonder3D~\cite{long2024wonder3d}. While 2D cross-view attention effectively enforces consistency across different views, it lacks explicit 3D spatial awareness, which can still cause incorrect shape generation (see Sec.\ref{sec:ablation_depthldm} in the supplementary material). For that reason, to enhance the model’s overall perception of 3D structures, we introduce an additional point-align attention mechanism. 

Specifically, in addition to using the latent partial representation $z_{mv}$ from the VAE-encoded multi-view incomplete depth image $I_{in}$, which serves as the primary condition concatenated with the input noise $x_{mv}$, we apply a point-align cross attention to incorporate the spatial feature $f_c$ of the partial point cloud as the additional condition. This allows us to generate semantic consistency and reliable multi-view latents $\hat{z} = \{z_{1}, z_{2}, ..., z_{m}\}$, where $m$ is the number of generated views. 
To extract the 3d structural point features, we use a pretrained Point-Bert ~\cite{yu2022point} as the point encoder to sample the points and extract patch-wise feature tokens, and concatenate the 2D coordinates of each point relative to every 2D view to align 2D and 3D domain, further form structural condition $f_c$. This additional condition, which encodes high-dimensional structural information, enhances the DepthLDM's perception of 3D space through the cross attention mechanism, thereby further improving the object structural consistency.

Finally, completed multi-view depth images can be obtained from those latents through VAE decoding, and the UNet $\epsilon_\theta$ is trained as follows:
\begin{equation}
\mathcal{L}_{MVD} = \mathbb{E}_{x_0, \epsilon, t} \Big[ \lVert \epsilon_{mv} - \epsilon_\theta(x_{mv}, z_{c}, t, f_c) \rVert_2^2 \Big],
\end{equation}
where $x_{mv}$ is the noisy sample of $z_{mv}$, $t$ is the time steps and $f_c$ is the point perception feature from Point-Bert. 

\subsection{Refinement Stage}
\label{Refinement}
In this stage, we aim to refine coarse point clouds formed from the initial stage. As illustrated in the bottom of Figure~\ref{fig:framework}, it begins by using a Point Denoising Network to remove the abnormal points introduced from the first stage. Then, an association relation construction module is utilized to learn a local transfer association from the reliable input structure to the corresponding coarse prediction. Finally, we stack two Association-Aware Point Upsamplers that propagate this association feature to the entire initial point cloud, further obtain the final complete point cloud.

\subsubsection{Point Denoising Network} 
Inevitable, the UNet tends to generate erroneous values in regions with large gaps~\cite{ran2024towards, ke2024repurposing}. Therefore, at the object boundaries in the generated depth maps, DepthLDM introduces some outliers, which, after projection, manifest as noise that adversely affects the subsequent upsampling process. Therefore, the Point Denoising Network (PDNet) is designed to identify and remove those abnormal points that deviate significantly from the object boundaries. To achieve this, the network learns to estimate the minimum distance between each point and the ground truth point cloud.

Specifically, as shown in Fig.~\ref{fig:framework}, our PDNet consists of an UNet-based feature extractor built by PointNet++ ~\cite{qi2017pointnet++} and an MLP-based prediction head. Given a coarse point cloud $P_c \in \mathbb{R}^{N_c \times 3}$, our feature extractor utilizes a multi-scale mechanism to compute the point's position feature $F_c \in \mathbb{R}^{N_c \times C_c}$ by considering the global coordinate distribution. Then, the prediction head is employed to directly predict the distance score $d_{pre} \in \mathbb{R}^{N_c \times 1}$ for each point. 

We use the single-sided chamfer distance between the coarse point cloud and the ground truth point cloud as the ground truth distance. This value serves as the supervisory signal for training the PDNet, which can be formulated as:
\begin{equation}
    \mathcal{L}_{r} = \mathrm{MSE}(\mathcal{D_{\theta}}(P_c), D_{gt})
\end{equation}
where $\mathcal{D_{\theta}}$, $P_c$, and $D_{gt}$ are our learnable distance estimation network, input coarse point cloud generated by DepthLDM, and ground truth distance, respectively. After the training, the network can adaptively predict the distance value of each point from the coarse point clouds. 

Once the minimal distances $d_{pre}$ have predicted, we use the top-k algorithm to filter the abnormal points to get the denoised point cloud $P_d \in \mathbb{R}^{N_d \times 3}$. Finally, following previous completion methods~\cite{yan2022fbnet, xiang2021snowflakenet}, we further merge $P_d$ with the input partial point cloud $P_{in}$ and sample to $N_i$ points by the farthest point sampling (FPS), forming the final initial point cloud $P_{init} \in \mathbb{R}^{N_i \times 3}$.

\begin{figure}[t]
\centering
\scalebox{0.45}{
\includegraphics[width=\textwidth]{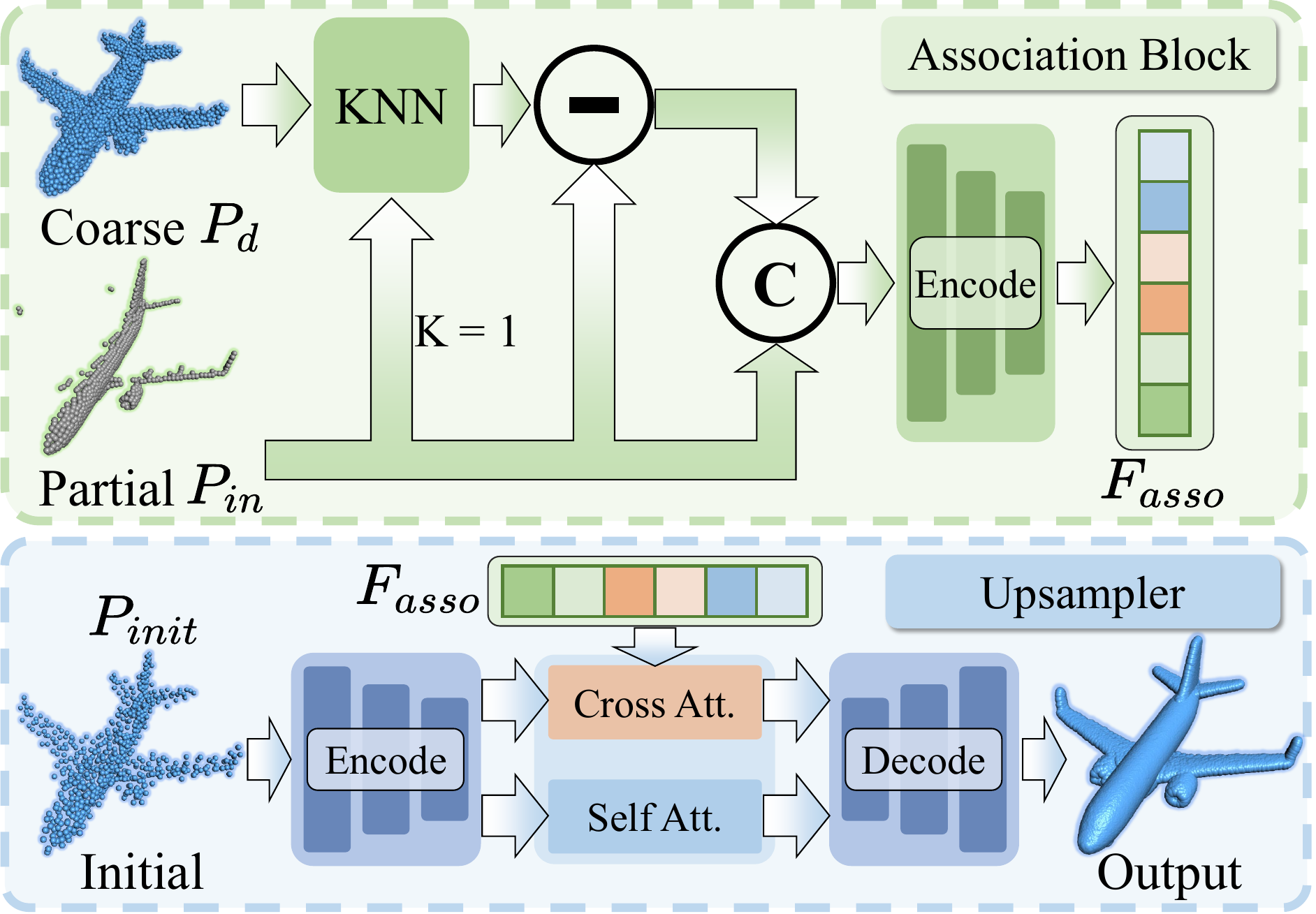}}
\caption{The overall structure of our Association Point Upsampler. "-" and "c" represent the subtraction and concatenate operation respectively.}
\label{fig:Upsampler}
\vspace{-2mm}
\end{figure}

\begin{figure*}[t]
\centering
\scalebox{0.90}{
\includegraphics[width=\textwidth]{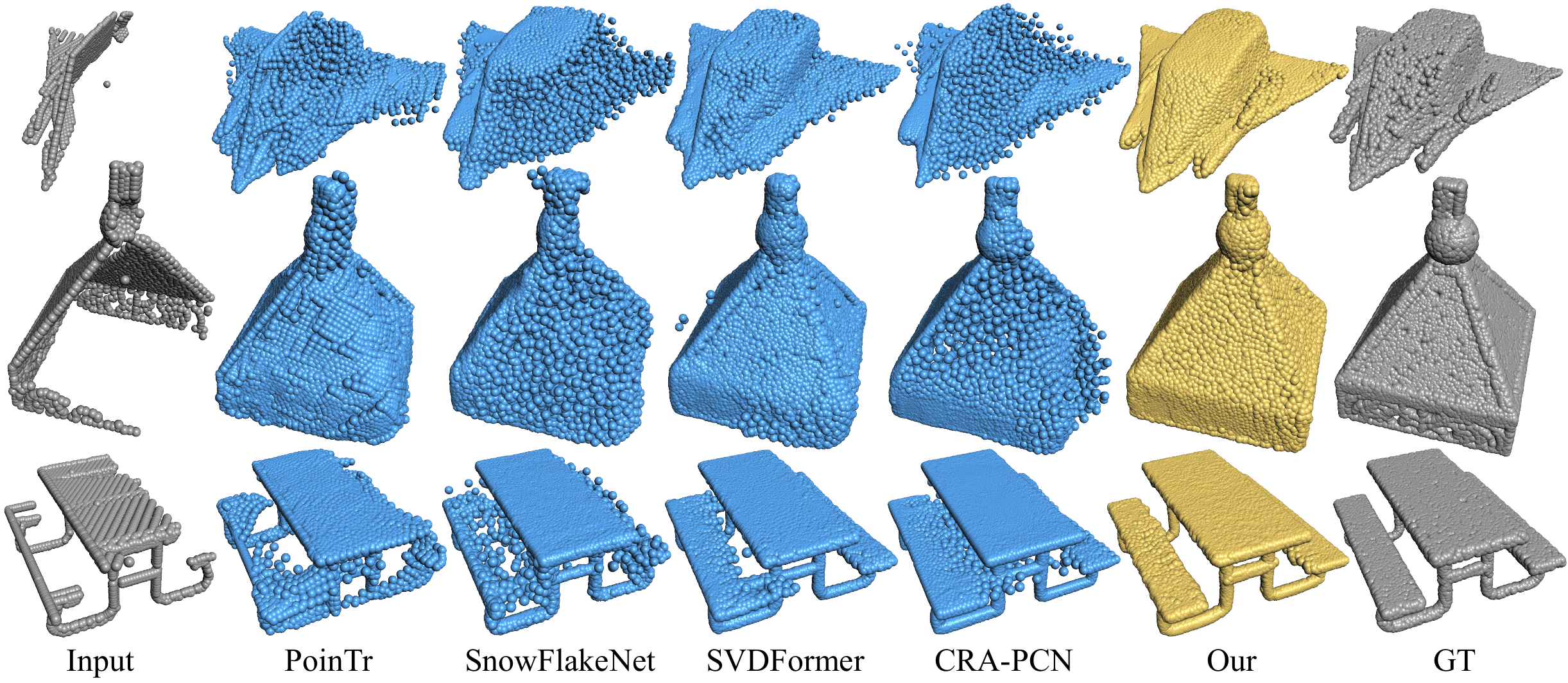}}
\caption{Visualization on the PCN's dataset. Our DiffPCN presents a better performance compared with previous methods.}
\label{fig:PCN}
\vspace{-2mm}
\end{figure*}

\begin{table*}[t]
\begin{center}
\scalebox{0.9}{
\begin{tabular}{l|cccccccc|cc}
\Xhline{2\arrayrulewidth}
Methods & Airplane & Cabinet &  Car  &  Chair & Lamp & Sofa & Table & Watercraft & $\text{CD}_{\text{avg}}$ ($\downarrow$) & F1 ($\uparrow$)  \\
\hline
PoinTr~\cite{yu2021pointr}  & 4.75 & 10.47 & 8.68 & 9.39 & 7.75 & 10.93 & 7.78 & 7.29 & 8.38 & -\\
SnowflakeNet ~\cite{xiang2021snowflakenet}  & 4.29 & {9.16} & 8.08 & 7.89 & 6.07 & 9.23 & 6.55 & 6.40 & 7.21 & 0.801\\
FBNet~\cite{yan2022fbnet} & {3.99} & {9.05} & {7.90} & {7.38} & {5.82} & {8.85} & {6.35} & {6.18} & {6.94} & - \\
SVDFormer~\cite{zhu2023svdformer} & 3.62 & 8.79 & 7.46 & 6.91 & 5.33 & 8.49 & 5.90 & 5.83 & 6.54 & 0.841 \\
PCDreamer~\cite{wei2024pcdreamer} & 3.51 & 8.72 & 6.89 & 6.71 & 5.64 & 8.32 & 6.24 & 5.84 & 6.49 & \underline{0.856} \\
CRA-PCN~\cite{Rong2024CRAPCNPC} & 3.59 & 8.70 & 7.50 & 6.70 & \underline{5.06} & 8.24 & 5.72 & 5.64 & 6.39 & - \\
PointCFormer~\cite{zhong2025pointcformer} & 3.53 & 8.73 & 7.32 & 6.68 & 5.12 & 8.34 & 5.86 & 5.74 & 6.41 & 0.855 \\
SymmCompletion~\cite{yan2025symmcompletion} & \underline{3.53} & \textbf{8.49} & \underline{7.30} & \underline{6.52} & \underline{5.06} & \underline{8.23} & \textbf{5.64} & \textbf{5.49} & \underline{6.28} & 0.853 \\ 
\hline
\textbf{DiffPCN} & \textbf{3.49} & \underline{8.52} & \textbf{7.26} & \textbf{6.51} & \textbf{4.90} & \textbf{8.15} & \underline{5.69} & \textbf{5.49} & \textbf{6.25} & \textbf{0.858} \\
\Xhline{2\arrayrulewidth}
\end{tabular}
}
\caption{Quantitative results in terms of $l1$ Chamfer Distance $\times 10^{3}$ (CD) and F1-Score@\%1 (F1) on PCN's dataset.}
\label{tab-pcn}
\vspace{-2mm}
\end{center}
\end{table*}

\subsubsection{Association Relation Construction.}
While the PDNet effectively removes abnormal points and provides a reliable initial point cloud, achieving high-resolution and structurally accurate reconstructions still requires further refinement. To address this, we initially introduce an association mechanism to extract the general transform feature.

As shown in Figure~\ref{fig:Upsampler}, given the partial input $P_{in}\in \mathbb{R}^{N_p \times 3}$ and the denoised coarse point cloud $P_{d} \in \mathbb{R}^{N_d \times 3}$, it begin with extracting a geometrically matched subset by a 1-nearest-neighbor lookup:
\begin{equation}
    P_{asso} = \mathrm{KNN}(P_{d}, P_{in}, k=1)
\label{eq:Association-Aware Partial Encoder}
\end{equation}
where we set k to 1 for getting unique neighbors, and $P_{asso}$ aligns exactly with its corresponding observed point in $P_{in}$. 

Since PointLDM inevitably introduces some reconstruction error, and $P_{in}$ is taken from the ground-truth coordinates, we compute the point-wise residual $\Delta P_{asso} = P_{in}- P_{d}$ as a measure of the correlation tendency between the reconstructed coordinates and the ground truth. We then employ an association encoder to obtain these association features that are used in the subsequent upsampling process.

Considering that these association features fundamentally learn a generalizable correction scheme, we cannot simply confine their encoding to either the coordinate domain or the feature domain. To address this, we propose a novel association encoder that operates in parallel, incorporating both a Point Transformer~\cite{zhao2021point} and a EdgeConv module~\cite{wang2019dynamic}. Detailed specifications of this encoder can be found in the Supplementary. 


\subsubsection{Association-Aware Point Upsampler}
\label{APU}



The Association-Aware Point Upsampler completes the refinement process by propagating the learned association features $F_{asso}$ into every point of the initial point cloud. It begins by encoding $P_{init}$ with a common point encoder which consists mainly of a point transformer and a self-attention block into high-dimensional features $F_m \in \mathbb{R}^{N_m \times C_m}$.

Next, we inject the local association feature via cross‐attention, using $F_m$ as queries and $F_{asso}$ as keys and values. To preserve residual detail from the initial point cloud, we simultaneously apply a parallel self‐attention branch to the original features $F_m$. We then concatenate these two streams along the channel dimension, and feed into a lightweight decoder $\mathcal{D}_p$ following SVDFormer~\cite{zhu2023svdformer}, which is composed of two self‐attention layers interleaved with MLPs. The whole module can be formulated as:
\begin{equation}
F_o = \mathcal{D}_p([\mathrm{SA}(F_m),\mathrm{CA}(F_m,F_{asso})]
\label{eq:Distribution Transfer Decoder}
\end{equation}
Finally, following the point-shuffle operation ~\cite{yang2018foldingnet}, we scatter the regressed offsets onto repeated copies of each coarse point, yielding the dense, high‐fidelity upsampled cloud.

\subsubsection{Loss Function}
For the refinement stage, we use the Chamfer Distance (CD) as the main distance function, which measures the average of the closest point distances between the output and the ground truth. The end-to-end loss function is formulated as:
\begin{equation}
    \mathcal{L}_{refine} = \mathcal{L}_{r}+\; \sum_{i=1}^{n} \mathcal{L}_{CD}\bigl(P_o^i, Q\bigr),
\end{equation}
where $\mathcal{L}_{r}$ corresponds to the PDNet loss, the second term is the total loss of our APU. Q is the ground truth point cloud, $P_o^i$ denote the final output of each APU, and $n$ is the stacked number of APU we used.


%% file: sec/5_experiment.tex
\section{Experiments}
In this section, we begin by introducing the datasets and evaluation metrics used for the point cloud completion tasks. Then, we compare our method with previous state-of-the-art approaches. Finally, we present an ablation study of our model.

\begin{table*}[t]
\begin{center}
\scalebox{0.95}{
\begin{tabular}{l|cc|cc|cc|cc}
\Xhline{2\arrayrulewidth}
\multirow{2}{*}{Method} & \multicolumn{2}{c|} {2048} & \multicolumn{2}{c|} {4096} & \multicolumn{2}{c|} {8192} & \multicolumn{2}{c} {16384} \\ \cline{2-9}
& CD ($\downarrow$) & F1 ($\uparrow$) & CD ($\downarrow$) & F1 ($\uparrow$) & CD ($\downarrow$) & F1 ($\uparrow$) & CD ($\downarrow$) & F1 ($\uparrow$) \\
\hline
PCN~\cite{yuan2018pcn}  & 9.77 & 0.320 & 7.96 & 0.458 & 6.99 & 0.563 & 6.02 & 0.638 \\
VRCNet~\cite{pan2021variational} & 5.96 & 0.499 & 4.70 & 0.636 & 3.64 & 0.727 & 3.12 & 0.791 \\
SnowflakeNet~\cite{xiang2021snowflakenet} & 5.71 & 0.503 & 4.40 & 0.661 & 3.48 & 0.743 & 2.73 & 0.796 \\
PDR~\cite{lyu2021conditional}& 5.66 & 0.499 & 4.26 & 0.649 & 3.35 & 0.754 & 2.61 & 0.817 \\
FBNet~\cite{yan2022fbnet} & {5.06} & {0.532} & {3.88} & {0.671} & {2.99} & {0.766} & {2.29} & {0.822} \\
AEDNet~\cite{fu2024aednet} & {5.12} & {0.522} & {3.75} & {0.675} & {2.90} & {0.770} & {2.24} & {0.832} \\
SymmCompletion~\cite{yan2025symmcompletion} & \textbf{4.89} & \underline{0.534} & \underline{3.65} & \textbf{0.691} & \underline{2.70} & \underline{0.782} & \underline{2.14} & \underline{0.850} \\
\hline
\textbf{DiffPCN} & \textbf{4.89} & \textbf{0.536} & \textbf{3.63} & \underline{0.690} & \textbf{2.68} & \textbf{0.791} & \textbf{2.06} & \textbf{0.857} \\
\Xhline{2\arrayrulewidth}
\end{tabular}
}
\caption{Quantitative results in terms of L2 Chamfer Distance $\times 10^{4}$ (CD) and F1-score@\%1 (F1) on the MVP dataset with different resolutions.}
\label{tab-mvp}
\end{center}
\vspace{-2mm}
\end{table*}

\subsection{Datasets and Evaluation Metrics}
{\bf PCN's dataset:}
PCN's dataset~\cite{yuan2018pcn} is a subset of ShapeNet. It consists of 28,974 training samples and 1,200 test samples from 8 categories, which is widely used as a benchmark for 3D point cloud completion.
{\bf MVP dataset:}
The MVP dataset~\cite{pan2021variational} includes 4,000 samples across 16 categories. Each model is virtually scanned from 26 camera positions, simulating realistic multi-view observations and better reflecting real-world variations in perspective. 
{\bf KITTI:}
KITTI~\cite{geiger2013vision} is a real-world dataset consist of 2,400 LiDAR-scanned car point clouds provided only as a test set, without fine registration. As models must be trained on other datasets and evaluated directly on these unaligned scans, KITTI poses a rigorous benchmark for assessing generalization and robustness in practical applications.

{\bf Evaluation Metrics:}
Similar to the previous methods~\cite{yuan2018pcn, yu2021pointr, xiang2021snowflakenet}, we adopt Chamfer distance (CD) and F1 score as our main metrics. For KITTI dataset, we use Minimal Matching Distance (MMD) and Fidelity (FD) to evaluate the performance.

\begin{figure}[t]
\centering
\scalebox{0.45}{
\includegraphics[width=\textwidth]{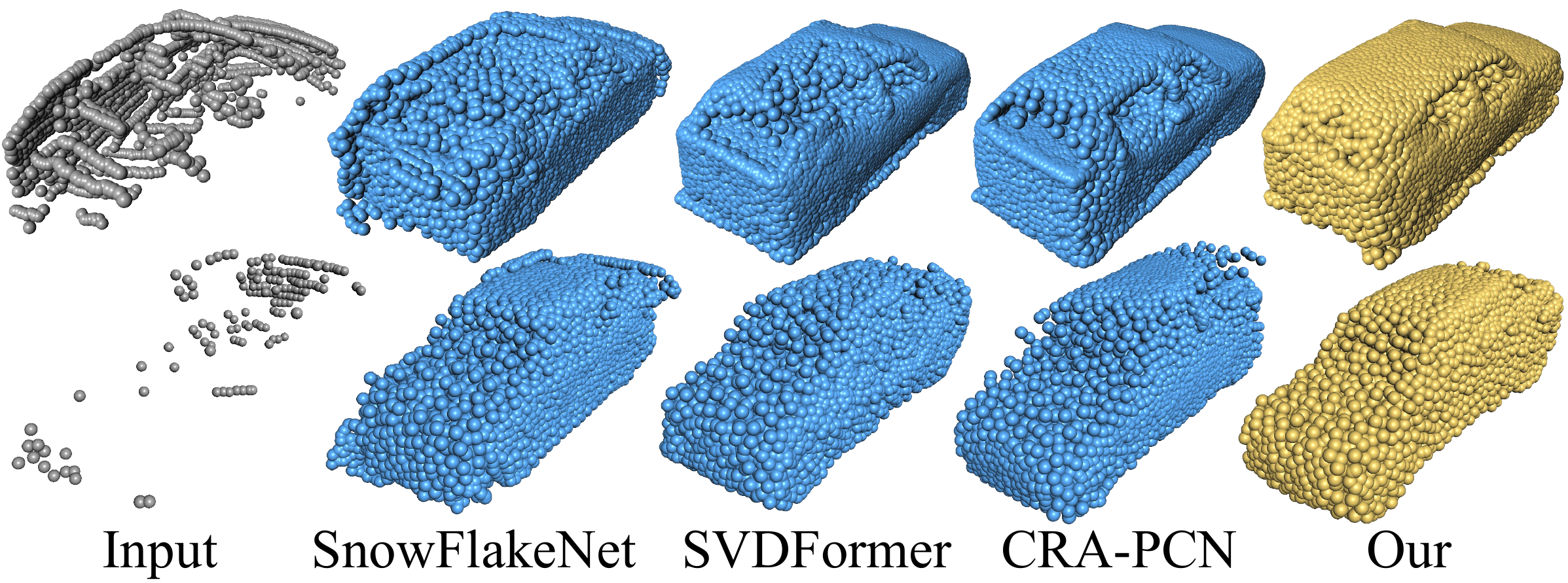}}
\caption{Visual comparisons on the KITTI dataset.}
\label{fig:KITTI}
\vspace{-5mm}
\end{figure}


\subsection{Experiment Results}
Initially, We evaluate our performance on the PCN's dataset. Table \ref{tab-pcn} and \ref{fig:PCN} shows that DiffPCN establishes new state-of-the-art results on PCN, achieving the lowest CD and highest F1. Qualitatively, the airplane in the first row illustrates how APU extends observed wing surfaces into unobserved regions, producing a smooth, high-fidelity reconstruction. The lamp in the second row demonstrates that PDNet effectively removes spurious points to yield a clean, noise-free completion. Finally, the bench in the third row showcases DepthLDM’s multi-view understanding, reconstructing intricate structures with precise geometric detail. These results demonstrate that DiffPCN, benefiting from the proposed modules, achieves more realistic and clear completions, ultimately reaching state-of-the-art performance. We further evaluate our model on the MVP dataset across all resolutions (2K, 4K, 8K, and 16K). The quantitative results, shown in Table~\ref{tab-mvp}, indicate that our method consistently achieves the best performance under different densities, which effectively shown the robustness of our diffusion-based method. 



\begin{figure}[t]
\centering
\scalebox{0.47}{
\includegraphics[width=\textwidth]{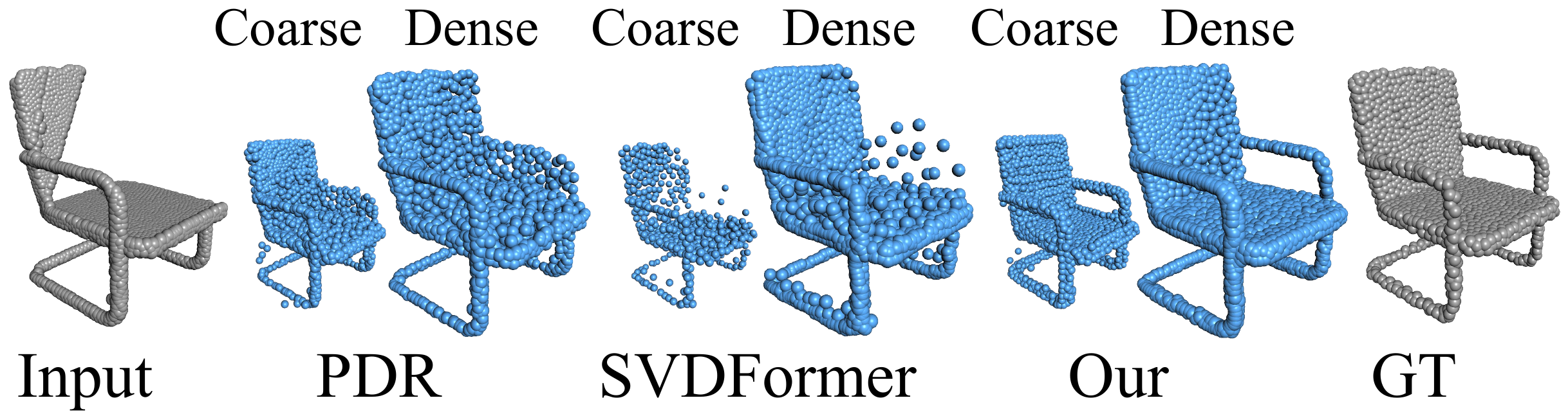}}
\caption{Visual comparisons of initial stage on MVP2K dataset.}
\label{fig:DepthLDM}
\vspace{-2mm}
\end{figure}


Finally, we test our model on the KITTI dataset. Following previous methods~\cite{zhu2023svdformer, Rong2024CRAPCNPC}, we use the pre-trained model trained on the PCN dataset for evaluation. The quantitative comparison is presented below:
\\[5pt]
\centerline{
\begin{tabular}{c|cccc}
\Xhline{2\arrayrulewidth}
 {Methods} & SVDFormer  & CRA-PCN & Symm & DiffPCN \\
\hline
FD ($\downarrow$) & 1.28 & 1.96 & 2.54 & \textbf{1.09} \\
MMD ($\downarrow$) & 0.75 & 1.08 & 1.72 & \textbf{0.61} \\
\Xhline{2\arrayrulewidth}
\end{tabular}
}
\\[5pt]
Our method achieves the best results in both MMD and FD, surpassing all previous state-of-the-art methods. This strong performance stems from the high generalization capability of our diffusion‐based DepthLDM: even on unprocessed LiDAR scans with variable point densities, DiffPCN produces reliable coarse predictions and robust refinements. Additionally, the qualitative results in Figure~\ref{fig:KITTI} demonstrate that our approach consistently achieves superior completeness. Notably, our model outperforms all existing methods across different scenarios, whether in relatively simple cases with denser input points or in more challenging cases with extremely sparse inputs.

\subsection{Ablation Study}
{\bf The effectiveness of DepthLDM.}
To validate the effectiveness of the depth latent diffusion model for the initial stage, we compare our DepthLDM with previous methods. Specifically, within our framework, we replace DepthLDM with the coarse point cloud generation methods from these approaches, including point-based SVDFormer~\cite{zhu2023svdformer} and diffusion-based PDR~\cite{lyu2021conditional}.
Since PDR only provides a pre-trained model on the MVP dataset, we conduct this comparison using the MVP dataset. As shown in Table~\ref{tab:depthldm}, our DepthLDM achieves the best performance, demonstrating its ability to generate a high-quality initial point cloud. We also provide qualitative comparisons in Figure~\ref{fig:DepthLDM}. In the initial stage, PDR and SVDFormer struggle to generate high-quality complete initial point clouds, which further affects the final quality for the refinement stage. In contrast, our model generates a structurally accurate initial point cloud, ultimately achieving the best final results.

\subsubsection{The effectiveness of APU.}
To quantify the impact of our upsampler, we replace the full APU with other upsampler form sota methods and test on the PCN dataset. As Table~\ref{tab:ablation-APU} shows, compared to the classical refiner SPD~\cite{xiang2021snowflakenet}, our APU achieves a significant improvement in both the CD and F1 scores, confirming its superior reconstruction fidelity. In addition, compared to the SGD~\cite{zhu2023svdformer} that also introduces the features of partial input via a cross-attention layer, our Association-Transformer provides unique displacement trend information, ultimately achieving superior results. These results validate that our association‐aware feature encoding and fusion strategy are key to generating dense and accurate point clouds.

\begin{table}[t]
\centering
\begin{tabular}{c|cc}
\Xhline{2\arrayrulewidth}
Initial Generator &  CD ($\downarrow$) & F1 ($\uparrow$) \\
\hline
PDR~\cite{lyu2021conditional} & 5.21 & 0.519 \\
SVDFormer~\cite{zhu2023svdformer} & 5.68 & 0.498 \\
DepthLDM ({\bf Ours}) & {\bf 4.89} & {\bf 0.536} \\
\Xhline{2\arrayrulewidth}
\end{tabular}
\caption{Quantitative comparison between different coarse point cloud generation (initial stage) methods on MVP2K.}
\label{tab:depthldm}
\vspace{-2mm}
\end{table}

\begin{table}[t]
\centering
\begin{tabular}{c|cc}
\Xhline{2\arrayrulewidth}
Upsampler $F_k$ &  CD ($\downarrow$) & F1 ($\uparrow$) \\
\hline
SPD~\cite{xiang2021snowflakenet} & 6.96 & 0.811 \\
SDG~\cite{zhu2023svdformer} & 6.45 & 0.846 \\
APU ({\bf Ours}) & {\bf 6.25} & {\bf 0.858} \\
\Xhline{2\arrayrulewidth}
\end{tabular}
\caption{Quantitative comparison about our APU on PCN dataset~\cite{yuan2018pcn}}
\label{tab:ablation-APU}
\vspace{-2mm}
\end{table}

%% file: sec/6_conclution.tex
\section{Conclusion}
In this paper, we propose DiffPCN, a novel depth diffusion-based coarse-to-fine framework for point cloud completion. By transforming 3D unordered point cloud into a 2D structured depth image, our approach leverages DepthLDM to generate a completed multi-view depth image with more details and project it to form a coarse point cloud. To remove the outlier points in the coarse point cloud, we design a Point Denoising Network to predict the confidence score and leverage the Top-K algorithm to choose high-confidence points. Further on, we stack two Association-Aware Point Upsamplers to progressively refine initial point clouds by introducing the association features. Extensive experiments on benchmark datasets demonstrate that DiffPCN outperforms existing methods both in geometric accuracy and completeness, highlighting the effectiveness of leveraging 2D LDMs for high-quality 3D reconstruction.